\documentclass[runningheads]{llncs}
\usepackage{graphicx}
\usepackage{amsmath}
\usepackage{amssymb}
\begin{document}
\title{A Computational Theory for Life-Long Learning of Semantics}
%
%
\author{Peter Sutor Jr.\inst{1}\orcidID{0000-0002-9200-3748} \and
Douglas Summers-Stay\inst{2}\orcidID{0000-0002-9286-5612} \and
Yiannis Aloimonos\inst{1}}

\authorrunning{P. Sutor, D. Summers-Stay et al.}

\institute{University of Maryland, College Park MD 20742, USA\\ \email{psutor@umd.edu}, \email{yiannis@cs.umd.edu} \and
U.S. Army Research Laboratory Adelphi, Adelphi MD 20783, USA\\
\email{douglas.a.summers-stay.civ@mail.mil}}

\maketitle              
\begin{abstract}
Semantic vectors are learned from data to express semantic relationships between elements of information, for the purpose of solving and informing downstream tasks. Other models exist that learn to map and classify supervised data. However, the two worlds of learning rarely interact to inform one another dynamically, whether across types of data or levels of semantics, in order to form a unified model. We explore the research problem of learning these vectors and propose a framework for learning the semantics of knowledge incrementally and online, across multiple mediums of data, via binary vectors. We discuss the aspects of this framework to spur future research on this approach and problem.

\keywords{Semantic Vectors  \and Hyperdimensional Computing \and Knowledge Representation \and Incremental Learning \and Dynamic Systems.}
\end{abstract}
\section{Introduction}

Semantic vector learning finds vector representations of semantic relationships observed in data that have useful properties, such as exhibiting similarity in the components of vectors via closeness under a metric, or allowing vector arithmetic to propagate semantic meaning. These vectors are learned by statistical distributions of co-occurrence in data (usually unsupervised), whose structure is embedded in a high dimensional space. With recent advances in neural networks, progress on semantic vectors has seen much success. But such models feel disjoint as semantic insight transfers poorly across them. State-of-the-art techniques rely on static datasets that estimate mapping functions across their distributions. Adapting models to another domain often requires complete retraining, or at least fine-tuning/transfer learning~\cite{featuretransfer}. In this paper, we examine the problem of learning semantics from supervised and unsupervised data to facilitate incremental and life-long learning. This is desirable as it synthesizes the semantics from not just multiple models, but potentially entirely separate domains (vision, linguistics, audio, etc.) into consistent vector representations for use in other tasks. The incremental process allows new models to come into existence at any time. We describe a general theoretical framework that can compute such vector representations in an online and perpetual way.

\section{Background Information}

\subsection{Related Work}

Our primary difference is the online and incremental learning capabilities of the vectors themselves, without a need for existing vector models, and the use of binary vectors to allow for useful and general data structure representation, as suggested by Kanerva~\cite{kanervahyper}. The problem of learning symbolic vectors for structured data has been explored using neural networks by Bordes et al.~\cite{knowledgembed}, with newer techniques~\cite{mlfromknowledge} learning from knowledge graph relationships, such as Freitas' DRNs~\cite{disrelnets}. Additionally, the reasoning, inference and lookup structures for embedded knowledge graphs are a well studied topic~\cite{inferencefreitas,freitasnlp}. Most are limited to static information, however there has been much work on the problem of completing relationships in knowledge graphs, such as \texttt{TransE}~\cite{transe} and its derivatives.

\subsection{Motivation}

The \texttt{word2vec}~\cite{w2v} paradigm forms semantic vectors for words by predicting the context around a word via an unsupervised process, given a corpus. The famous $king - man + woman \approx queen$ example shows its ability to learn deep analogies between word vectors. However, consider: $smokestack - cigarette + firework$. It's clear that something like ``missile'' would be a valid answer to this analogy. This is not the case for \texttt{word2vec}, which does not return ``missile'', or anything sensible, for even the top 100 matches for the popular Google News \texttt{word2vec} model~\cite{googlenews}. Why does it fail for this example, but not for much deeper analogies such as $death - life + good \approx bad$? The analogy is purely visual/functional here. However, \texttt{word2vec} never `sees' anything visual, only patterns in words, where such a relationship is unlikely to occur. A similar situation is apparent in the auditory domain: $quack - bird + car = honk$ also fails in \texttt{word2vec}.

Clearly, learning human-level semantics requires integration of hierarchically built-up data and relationships from differing domains. For example, neural networks generalize better by using character level patterns as well as word level~\cite{kim2016character,zhang2015character}. A general model should include multiple forms and levels of perception into a single semantic model that takes everything into account, from a general intelligence perspective. This is supported in the biological setting, where it seems that neurons can take on other roles over time, when necessary~\cite{blindlocalization}, suggesting that many cortical neurons treats information in a similar way.

\subsection{Incremental, Online, and Generalized Semantic Learning}

Our goal is to take raw input from various perceptions, outputs of other models and unify them. Such models give semantically significant relationships, that could be learned or heuristic in nature. This could be between sequences of raw data, more complex mappings of words to parts of speech, or dependency arcs between words. In the visual domain, patterns of pixels can map to classification labels. We can visualize this as building a graph of relationships. Short of numeric regression, most outputs of models can be expressed as edges between nodes, due to their discrete nature. New info is easily absorbed by simply updating the graph and statistics on it. We wish to find appropriate vectors for the nodes. Since new knowledge comes in an online process, as new observations, new nodes or edges in the graph, or the addition of new models to the system, these vectors themselves must be quickly adjusted in an equally online and incremental process.

\subsection{Long Binary Vectors as General Features}

As noted by Kanerva~\cite{kanervahyper}, long binary vectors, on the order of 10,000 components, are a promising vector representation. Consider the space $\mathbb{B}^n = \{0, 1\}^n$. This space contains $|\mathbb{B}^n| = 2^n$ possible vectors. No computational system needs anywhere near $2^{10,000}$ vector representations. This space represents the corners of a 10,000 dimensional hypercube, so every point in the space has the same distribution of distances to other points. Let:

\begin{equation}\label{eqn:hamming}
H(x,y) = \sum_{i=0}^{n}{\mathbb{I}_{x_i \neq y_i}(i)} = \sum_{i=0}^{n}{x_i \oplus y_i} = |x \oplus y|
\end{equation}

\noindent be the Hamming Distance, where $\mathbb{I}_{x_i \neq y_i}$ is the indicator that the bits at $i$ between $x$ and $y$ disagree, returning 1 in this case, and 0 when they agree, $a \oplus b$ is the bitwise exclusive-or (XOR) operation, and $|a|$ is the number of 1's in $a$. The number of bits of disagreement is the distance between points. Let:

\begin{equation}\label{eqn:nhamming}
H_N(x,y) = \frac{1}{n}H(x,y)
\end{equation}

\noindent be the Normalized Hamming Distance, which expresses the distance on a real scale of 0 to 1. Assuming each possible vector is equally likely, the average Normalized Hamming Distance is 0.5. Furthermore, under these assumptions, with $n = 10,000$, their distribution is binary with mean 5000 bits and standard deviation 50; this implies that for any significant deviation from distance 0.5, the distribution quickly becomes very sparse. An astronomically vast majority of vectors have distances very close to 0.5. Randomly drawn vectors are nearly guaranteed to differ from one another by about 5000 bits, or distance 0.5. 

This distribution is resistant to noise in its vectors, as a large portion of the bits in any vector would have to be randomly flipped before the distance becomes very big between another nearby vector. Two related vectors differing even by 5\% of their bits is so astronomically unlikely that they may as well still be the same vector. As pointed out by Kanerva, such long binary vectors also have the property of encoding various forms of information that can be later recovered even under noisy conditions. This comes primarily from three operations:

\begin{enumerate}
\item The XOR $c = a \oplus b$: Since XOR is an involution when one operator is fixed, and associative and commutative, $c \oplus a = (a \oplus b) \oplus a = a$. We can exactly recover $a$ or $b$ if we have one or the other, or approximately with noise.
\item The permutation $\Pi$: This permutes a vector $x$'s components into a new order, by computing the product $\Pi x$. If the permutation is randomly generated for a long binary vector, the new binary vector is very likely to have a distance~(\ref{eqn:nhamming}) near $0.5$. We can represent $\Pi$ as a permutation of index locations $1$ to $n$. The product simply swaps components of $x$ to the order in $\Pi$.
\item The consensus sum, $+_c(A)$, over the set of vectors $A$: This sum counts 1's and 0's component-wise across each element of $A$, and sets the component to the corresponding value with the bigger count. Ties, only possible in a sum of an even number of elements, can be broken by randomly choosing.
\end{enumerate}

\noindent Note that mapping by XOR or permuting preserves distances. For mapping $a$:

\begin{equation}\label{eqn:xordistance}
\begin{array}{cc}
H(a \oplus x, a \oplus y) = |a \oplus x \oplus a \oplus y| = |a \oplus a \oplus x \oplus y| = |x \oplus y| \\
H(\Pi x, \Pi y) = |\Pi x \oplus \Pi y| = |\Pi (x \oplus y) = |x \oplus y|
\end{array}
\end{equation}

\noindent as permutation is distributive, thus $\Pi a \oplus \Pi b = \Pi (a \oplus b)$, and permuting doesn't change the number of 1's or 0's, so $|\Pi c| = |c|$, for any $a$, $b$, and $c$. 

\subsection{Representing Data Structures With Binary Vectors}

We can create binary vector abstractions of simple data structures:

\vspace{2mm}
\noindent \textbf{Sets:} A set of data $\{\zeta_1, \zeta_2, ..., \zeta_m\}$, given a mapping between $\zeta_i$ and binary vectors $\{z_1, z_2, ..., z_m\}$, can be represented as $z = z_1 \oplus z_2 \oplus ... \oplus z_m$. A union of sets $x$ and $y$, of $m_1$ and $m_2$ elements, with no elements in common, is:

\begin{equation}\label{eqn:union}
x \cup y = x \oplus y = x_1 \oplus x_2 \oplus ... \oplus x_{m_1} \oplus y_1 \oplus y_2 \oplus ... \oplus y_{m_2}
\end{equation}

\noindent However, equation~(\ref{eqn:union}) is not the general case. With unrestricted $x$ and $y$:

\begin{equation}\label{eqn:setdiff}
x \oplus y = (x - y) \cup (y - x)
\end{equation}

\noindent Furthermore, set intersection and complement is impossible to compute without knowing the original vector values of the components.

\vspace{2mm}
\noindent \textbf{Ordered Pairs:} Represented by tuple $\zeta_x = (\zeta_x, \zeta_y)$, where $\zeta_x$, $\zeta_y$ and $\zeta_z$ are data points. If mapped to binary vector representations $x$, $y$, and $z$:

\begin{equation}\label{eqn:orderedpair}
z = \Pi x \oplus y \textrm{ , or } z = +_c(\{\Pi x, y\})
\end{equation}

\noindent This random permutation $\Pi$ then denotes the data type of $\zeta_z$.

\vspace{2mm}
\noindent \textbf{Sequences:} We can interpret a sequence $\zeta_z = \zeta_{z_1} \zeta_{z_2 ... \zeta_{z_m}}$ of a particular data type as a 2-tuple $\zeta_z = (\zeta_{z_1} ... \zeta_{z_{m-1}}, \zeta_{z_m})$. With binary vectors $z_1, z_2, ..., z_m$, this reduces to a succession of pairings via equation~(\ref{eqn:orderedpair}):

\begin{equation}\label{eqn:sequences}
z = \Pi^{m-1}z_1 \oplus \Pi^{m-2}z_2 \oplus ... \oplus \Pi^{m-i}z_i \oplus ... \oplus \Pi z_{m-1} \oplus z_m
\end{equation}

\noindent where $\Pi^j$ is a permutation $\Pi$ that permutes itself $j$ times. If $\Pi$ is random, then $\Pi^j$ appears random too, with an expected distance from $\Pi$ near 0.5. Equation~(\ref{eqn:sequences}) holds inductively as $\zeta_{z_1} \zeta_{z_2}$ is $\Pi z_1 \oplus z_2$, thus $\zeta_{z_1} \zeta_{z_2} \zeta_{z_3}$ is:

$$\Pi (\Pi z_1 \oplus z_2) \oplus z_3 = \Pi \Pi z_1 \oplus \Pi z_2 \oplus z_3 = \Pi^2 z_1 + \Pi z_2 + z_3$$

\noindent Continuing this for an arbitrarily long pattern gives equation~(\ref{eqn:sequences}). Similar reasoning will get us the equivalent with $+_c$ replacing XOR for the sum:

\begin{equation}\label{eqn:sequenceconsensus}
z = +_c(\{\Pi^{m-1}z_1, +_c(\{\Pi^{m-2}z_2, ... +_c(\{\Pi^{m-i}z_i, ...\}) ...\})\})
\end{equation}

\vspace{2mm}
\noindent \textbf{Data Records:} A unit of data containing one or more fields, where each component has a specific meaning. Let a data record of type $R = [r_1 r_2 ... r_m]$ of $m$ fields, where each binary vector $r_i$ is a field; for example, ``name'', ``age'', ``gender'', etc., for a record of a person. To set values to a field, we bind each field to a value. Since the XOR of two vector mappings represents a set - essentially a bound field - we can use $r_i \oplus v_i$ to bind a $v_i$ to $r_i$. Given $r_i$, we can recover $v_i$, or vice versa. We can generalize a bound data record $R_v$ by:

\begin{equation}\label{eqn:datarecord}
\begin{split}
R_v &= [r_1 r_2 ... r_m][v_1 v_2 ... v_m]^T\\ &= r_1 \oplus v_1 + r_2 \oplus v_2 + ... + r_m \oplus v_m = +_c(\{r_i \oplus v_i\})
\end{split}
\end{equation}

\noindent To isolate the value of a field $r_i$, we compute $R_v \oplus r_i$. The contribution of unrelated fields will generally create random noise when $r_i$ distributes across them, but only the contribution of the $r_i \oplus v_i$ will be non-random and significant, as $r_i \oplus r_i \oplus v_i = v_i$, generating a signal that is close to $v$. This is because each bit of $R_v$ represents the majority of the terms $r_i \oplus v_i$. High-dimensional binary vectors resist such noise, so the closest neighbor to $R_v \oplus r_i$ is likely $v_i$.

\section{Life-Long Learning of Semantics}

\subsection{A Geometric Interpretation of Semantics}

Consider the task of learning semantic vectors given a knowledge graph $K = (V, E)$ consisting of $n_v$ vertexes $V = \{v_1, ..., v_{n_v}\}$ and $n_e$ directed, weighted edges $E = \{e_1, ..., e_{n_e}\}$. Let any $v_i$ have \textit{mass} equal to the number of times its relationships have been observed, or a similar statistic. We can equate this to a simple, undamped spring-mass system, with a few caveats. Since it is directed, the target end of the edge is seen as fixed, converting mass (by edge weight) into constant acceleration towards the target, or a \textit{connecting force}. Let mass generate a repelling \textit{proximal force} similar to gravity, to prevent singularities. This is somewhat akin to self-organizing maps ~\cite{som}. Suppose vertexes exist in $n$-dimensional space accompanied by a distance metric, with locations randomly chosen. The structure of the knowledge graph causes the forces to be very high. We wish to minimize them by placing vertexes in a ``better'' location. In order to facilitate this, we define an ``anchoring'' vertex that cannot move from it's random position, connected to all vertexes with incoming edges. A minimized configuration of this system is equivalent to a good semantic placement of vectors, if an outgoing edge from a vertex signifies it should be more ``similar'' to the target. Efficient algorithms exist for this minimization in the real domain~\cite{liu2013fast}.

\subsection{Binary Vector Analogues}

We will now construct a binary vector analogue of the geometric, knowledge graph based semantic minimization problem. Let $K$ be of $m$ vertexes, $X^{(k)} \in \mathbb{B}^{m \times n}$ be a binary matrix, where $m$ rows correspond to positional vectors for $m$ vertexes in $K$, $n$ is the number of dimensions, and $k$ is the iteration step. Initially, $X^{(0)}$ is randomly selected. Let row 1 be the anchoring vector. Our problem statement is to find a perturbation matrix $X \in \mathbb{B}^{m \times n}$ such that:

\begin{equation}\label{eqn:tensionmin}
\textrm{arg}\min_X(T(X^{(k)} + X))
\end{equation}

\noindent where $T$ denotes the \textit{total tension} in our system. Then, we set $X^{(k+1)} = X^{(k)} + X$ and continue. This is the sum of unresolved forces across all bits of a given $A$:

\begin{equation}\label{eqn:tensiondef}
T(A) = \sum_{i = 1}^{m}{\sum_{j = 1}^{n}{\max(F_{conn}(A, i, j) + F_{prox}(A, i, j), 0)}}
\end{equation}

\noindent where $F_{conn}$ and $F_{prox}$ are the connective and proximal forces for vector $i$, bit $j$ in $A$. If the sum of these two forces for a bit are positive, it means the bit wants to change, otherwise it wants to remain the same. Thus, a system with all negative or 0 resultant forces is considered minimized. The proximal force is:

\begin{equation}\label{eqn:proximal}
F_{prox}(A, i, j) = \sum_{k = 1, k \neq i}^{m}{\dfrac{M_iM_k}{H(A_i, A_k)^2} C_{prox}(A_{ij}, A_{kj})}
\end{equation}

\noindent where $M_i$ and $M_k$ are the masses of the corresponding rows $A_i$ and $A_k$ of $A$. This force is clearly analogous to gravitational force between two masses, but a repulsive one, with normalized Hamming distance between them. Likewise:

\begin{equation}\label{eqn:connective}
F_{conn}(A, i, j) = \sum_{k = 1, k \neq i}^{m}{M_iW_{ik}C_{conn}(A_{ij}, A_{kj})}
\end{equation}

\noindent is the connective force, where $W_{ik}$ denotes the directed edge weight between vectors $i$ and $k$, which is non-zero and less than or equal to 1 if it exists, and 0 otherwise. The functions $C_{prox}$ and $C_{conn}$ are special functions defined by:

\begin{equation}\label{eqn:selector}
\begin{array}{cc}
    C_{prox}(a, b) = \left\{\begin{array}{rr}
   	1, & \text{if } a = b\\
    -1, & \text{if } a \neq b
   	\end{array}\right\} & \hspace{3mm}
    C_{conn}(a, b) = \left\{\begin{array}{rr}
   	-1, & \text{if } a = b\\
    1, & \text{if } a \neq b
   	\end{array}\right\}
\end{array}
\end{equation}

\noindent which decide the direction forces act on the bit, depending on if the bits connected have the same or differing value. As proximal force should push $a$ away from $b$'s value, it will add tension to the system only if the bits aren't as far away as possible. Connective force works the other way around, so $C_{conn} = -C_{prox}$. Consider the sum of connective and proximal forces from equations~(\ref{eqn:connective}) and~(\ref{eqn:proximal}):

\begin{equation}\label{eqn:forcesum}
\begin{split}
F & = \sum_{k}{M_i W_{ik}C_{conn}(A_{ij}, A_{kj})} + \sum_{k}{\dfrac{M_iM_k}{H_N(A_i, A_k)^2} C_{prox}(A_{ij}, A_{kj})}\\ & = \sum_{k = 1, k \neq i}^{m}{M_iC_{conn}(A_{ij}, A_{kj}) \left[W_{ik} - \dfrac{M_k}{H(A_i, A_k)^2}\right]}
\end{split}
\end{equation}

\noindent by substitution from~(\ref{eqn:selector}). Thus, our total tension function to minimize is:

\begin{equation}\label{eqn:totaltensionfinal}
T(A) = \sum_{i = 1}^{m}{\sum_{j = 1}^{n}{\max\left(\sum_{k}{M_iC_{conn}(A_{ij}, A_{kj}) \left[W_{ik} - \dfrac{M_k}{H(A_i, A_k)^2}\right]}, 0\right) }}
\end{equation}

\begin{figure}[t]
\includegraphics[width=\textwidth]{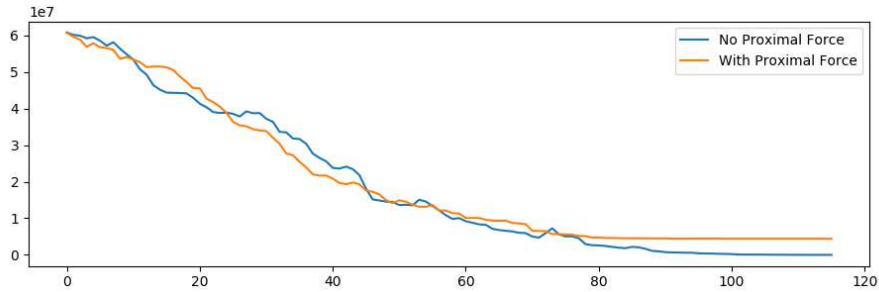}
\caption{Example minimization per random row of a randomly connected 50 node graph's binary vectors via the greedy method. Without proximal force it reaches 0.} \label{fig:binary}
\end{figure}

\noindent We can see from~(\ref{eqn:totaltensionfinal}) that~(\ref{eqn:forcesum}) is the term to minimize across $i$ and $j$ to find perturbation $X$ in~(\ref{eqn:tensionmin}). This can be found na\"{i}vely and greedily by randomly trying to flip bits with row-weighted probabilities from~(\ref{eqn:forcesum}) for a particular vector and observing the change in total tension~(\ref{eqn:totaltensionfinal}). Suppose you flip the first bit in a row that satisfies an adaptive threshold for the ratio of the row's tension it accounts for. The Many-Body problem prevents feasible computation of all Hamming Distances. We make the simplifying assumption that proximal force exists only for connected nodes. Then, one can efficiently compute the difference in forces for that row, and even the energies for all other rows affected, using simple $\pm 1$ of Hamming Distances. After flipping, the next bit in the row which gives a sufficient negative difference is found. The process repeats until no such bits can be found and a new row is randomly selected. Figure~(\ref{fig:binary}) shows the minimization of this technique for a randomly connected system. When proximal force is ignored, minimization to 0 is guaranteed. But unwanted singularities can occur on connected components, which proximal force avoids, although a minimum of 0 is not guaranteed. Minimization grows linearly with the number of nodes. This technique can be combined with Simulated Annealing and (self) supervised Q-Learning over rewards of minimization of~(\ref{eqn:totaltensionfinal}), which can not only be online but also learn \textit{how} to minimize.

\begin{figure}[t]
\includegraphics[width=\textwidth]{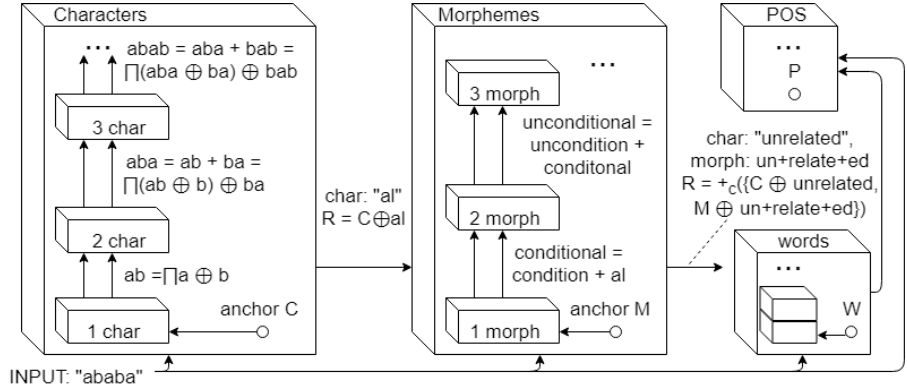}
\caption{A pipeline of knowledge graph computation of vertexes for a simple linguistic examples: raw characters, morphemes, words and parts of speech. Sequences of a space are embedded within by permuting the new element in the sequence with a static $\Pi$ randomly chosen by the space. Crossing semantic spaces is performed by a consensus record on components of the relationship, where anchor vertexes are the fields.} \label{fig:l3spaces}
\end{figure}

\subsection{Incremental Life-Long Learning of Semantics}

We now propose a general model for incremental, online learning of semantic binary vectors. Initially, suppose we have some form of raw inputs of data (unsupervised), whose possible values are discrete. For a working example, as shown in Figure~(\ref{fig:l3spaces}), let this be the space of characters in text. As input comes in to the system, we start off by letting each new character or character sequence have its own vertex and random binary vector (on order of $n = 10,000$ components). The space of characters is represented by a random, static vector $C$. As text is read as raw input, the system builds connections between characters by placing directed edges when two characters are next to each other. The weights of the edges are the probability of the transitioning occurring over the number of occurrences of the character. The number of observations of the characters become its ``mass''. Simultaneously, when the system is presented at least a sequence of 2 characters, it can form a binary vector representation for this via a random and static permutation $\Pi$, and~(\ref{eqn:sequences}). From the perspective of single characters, this is the position of sequences of 2 characters. Generally, for sequences of $l \geq 3$:

\begin{equation}\label{eqn:seqmapper}
\zeta_{x_1} \zeta_{x_2} ... \zeta_{x_l} = \zeta_{x_1} ... \zeta_{x_{l-1}} + \zeta_{x_2} ... \zeta_{x_l} \rightarrow \Pi(x_1 ... x_{l-1} \oplus x_2 ... x_{l-1}) \oplus x_2 ... x_l
\end{equation}

\noindent is the position of an $l$ sequence from the perspective of an $l-1$ sequence.

More generally, in a growing knowledge graph, this is two directed arcs $a$ and $b$ coming together to a new vertex $ab$ for the sequence. However, in the space of a sequence, co-occurrences between other similar length sequences can be recorded, and the tension in this system minimized for new binary vectors, to get a stronger representation. Here, $ab$ is represented by a local vector $c$. Since we can compute a mapping between $ab$ and $c$ by $ab \oplus c$, subsequently bigger sequences should use this mapping. Frequencies from data and new edges in the knowledge graph are recorded this way until the model decides to minimize tension in each subsystem for new data. This process can be performed across other models on the data, as Figure~(\ref{fig:l3spaces}) shows. To map between data types, we compute a record $R$ that binds each value to types as fields with~(\ref{eqn:datarecord}). We can do this across different forms of raw data as well, as long as we have a model that will map other data types to raw data (such as a classifier or static model). This is shown for morphemes, words, and parts of speech in the figure, but this can be done from the space of images to linguistic data, or other forms of raw data.

Suppose a new, supervised model appears. Since mapping between data types is performed by records, we can easily add new information to it, as consensus sum is commutative. This is tolerant to incremental learning. Since the dynamic system for learning binary vectors proceeds arbitrarily, and edges/vertexes can be added to $K$ quickly, this is also online and incremental. We can continue to add new models, read new data, and minimize tension in each system. If a model exists that adds edges across different data types, we can cross boundaries semantically and form more complicated data records. These can form hierarchical structures of semantic organization across many mediums of data.

\section{Discussion and Conclusion}

\noindent \textbf{Semantic Vector Operations:} Local Hamming Distance measures similarity for occupants of the same hierarchy, or from the perspective of a smaller sequence to a larger sequence. However, similarity across entire semantic spaces is a difficult, potential research area. Data records can isolate values of fields and compare their distances. We can also map ALL vertexes to the perspective of a space of concepts that all data belongs in. Since vectors start off random, it's possible to measure the likelihood of two vectors to be within a certain distance, as Kanerva~\cite{kanervahyper} describes. XOR can also construct and deconstruct complex semantic concepts and get vectors for new data, or even unbound anchor points.

\vspace{2mm}
\noindent \textbf{Self Improvement:} Self benchmarking of semantic learning can be performed. If the nearest known neighbor for a data record with some missing fields is incorrect, the system has can leverage the knowledge graph and upscale the weights such that minimization of~(\ref{eqn:totaltensionfinal}) makes it correct. So, if it fails often in one area, it can over-represent similar tests in the future, allowing targeted improvement. If incrementally trained enough, the system could form interesting questions for humans to answer, in order to learn relationships that are not easy to model. For example, if vectors are close, but share no knowledge graph edges, the system can predict the odds of this randomly happening to either be sure a relationship exists, or file it as a candidate relationship to cross-reference later. Thus, this lends itself to interacting with humans and asking useful questions.

\vspace{2mm}
\noindent \textbf{Conclusion:} We have proposed a novel, general theory for directly learning distributed binary semantic features from arbitrary data that can be put into the form of a knowledge graph, whether supervised or unsupervised, to incrementally learn high dimensional, semantic binary vectors online. This life-long learning is a promising technique for combining semantic knowledge across many existing models and raw data to build deep, useful, hierarchical representations. In future work, we hope to implement and test this model on at least linguistic and visual data in a manner that enables empirical testing of its properties.

\section{Acknowledgements}
The support of ONR under award  N00014-17-1-2622 is gratefully acknowledged.

%
%

%
%
%
\bibliographystyle{splncs04}
\bibliography{l3}
%




\end{document}